\pdfoutput=1
\PassOptionsToPackage{table}{xcolor}
\documentclass[11pt]{article}

\usepackage[final]{acl}

\usepackage{times}
\usepackage{latexsym}
\usepackage[T1]{fontenc}
\usepackage[utf8]{inputenc}
\usepackage{amsmath}
\usepackage{microtype}
\usepackage{inconsolata}
\usepackage{multirow}
\usepackage{graphicx}
\usepackage{subcaption}
\usepackage{array}

\title{Can Deception Detection Go Deeper? \\ Dataset, Evaluation, and Benchmark for Deception Reasoning}

\author{Kang Chen$^{1,2*}$, Zheng Lian$^{1}$\thanks{Equal Contribution}$^,$\thanks{Corresponding Author}, Haiyang Sun$^{3}$, Rui Liu$^{4}$, Jiangyan Yi$^{1}$, Bin Liu$^{1}$, Jianhua Tao$^{5,6}$  \\
$^1$Institute of Automation, Chinese Academy of Sciences, $^2$Peking University, \\
$^3$Shanghai Jiao Tong University, $^4$Inner Mongolia University \\
$^5$Department of Automation, Tsinghua University \\
$^6$Beijing National Research Center for Information Science and Technology, Tsinghua University\\
\texttt{2201210163@stu.pku.edu.cn, lianzheng2016@ia.ac.cn}}

\begin{document}
	\maketitle
	
	\begin{abstract}
	Deception detection has attracted increasing attention due to its importance in real-world scenarios. Its main goal is to detect deceptive behaviors from multimodal clues such as gestures, facial expressions, prosody, etc. However, these bases are usually subjective and related to personal habits. Therefore, we extend deception detection to deception reasoning, further providing objective evidence to support subjective judgment. Specifically, we provide potential lies and basic facts and then analyze why this sentence may be a lie by combining factual inconsistencies and intent behind them. Compared with deception detection, this task is more applicable to real-world scenarios. For example, in interrogation, the police should judge whether a person is lying based on solid evidence. This paper presents our initial attempts at this task, including constructing a dataset and defining evaluation metrics. Meanwhile, this task can serve as a benchmark for evaluating the complex reasoning capability of large language models. \textcolor[rgb]{0.93,0.0,0.47}{Our code and data are provided in the supplementary material.}
	\end{abstract}

	\section{Introduction}
	Deception is defined as an intentional attempt to mislead others \cite{depaulo2003cues}. Detecting deceptive behaviors is challenging even for humans, generally requiring specialized knowledge. Despite its difficulties, deception detection is an important research topic due to its widespread applications, such as airport security screening, court trials, and personal credit risk assessment \cite{masip2017deception}.
	
	Deception detection aims to identify deceptive behavior from multimodal clues (such as blinking, stuttering, etc.). Current research mainly focuses on laboratory-controlled or in-the-wild scenarios \cite{karnati2021lienet, speth2021deception}. The former recruits subjects and triggers their deceptive behaviors in well-designed psychological paradigms \cite{abouelenien2016detecting}. However, some researchers question the practicality of laboratory-controlled datasets because they are different from real deceptive behaviors \cite{vrij2008detecting, fitzpatrick2022automatic, fornaciari2020fake}. Therefore, in recent years, researchers have paid more attention to real-life datasets \cite{csen2020multimodal}.
	
	However, such judgment is usually subjective and related to personal habits. In real applications, we need to provide solid evidence to support the judgment. Therefore, we extend deception detection and propose a new task called ``deception reasoning''. In this task, we provide a potential lie and basic facts and try to figure out why this sentence may be a lie by considering factual inconsistencies and the intent behind them.
	
	In this task, our main goal is not to improve the authenticity of deception but to focus on the reasonableness of reasoning. Therefore, to reduce the cost of data collection, we use GPT-4 to synthesize dialogues with deceptive behaviors. Besides datasets, we define four metrics to comprehensively evaluate the reasoning results: \emph{accuracy}, \emph{completeness}, \emph{logic}, and \emph{depth}. The main contributions of this paper are summarized as follows:
	\begin{itemize}
		\item We propose a new task, deception reasoning. Different from deception detection, we further provide objective evidence for potential lies.
		
		\item To facilitate subsequent research, we construct a dataset and evaluation metrics.
		
		\item This task can also serve as a benchmark to evaluate the complex reasoning capability of large language models (LLMs).
	\end{itemize}

	Section \ref{sec:related} reviews recent works. In Section \ref{sec:data}, we introduce our data generation pipeline. In Section \ref{sec:eval}, we define evaluation metrics and report the performance of various LLMs on deception reasoning. Finally, we conclude this paper in Section \ref{sec:conclusion}.

	\section{Related Works}
	\label{sec:related}
	In this section, we first review existing works on deception detection and LLMs. Since we focus on deception reasoning, we further review some works on evaluating reasoning capabilities.
	
	\subsection{Deception Detection}
	Deception detection aims to identify deceptive behavior based on multimodal clues. Current works in this field are mainly conducted in laboratory-controlled or in-the-wild scenarios.
	
	In laboratory-controlled setups, researchers often use well-designed psychological paradigms to induce deception. For example, \citet{derrick2010border} asked participants to commit mock crimes. They were rewarded if they could convince the professional interviewer of their innocence. \citet{perez2014multimodal} and \citet{abouelenien2016detecting} collected data using three scenarios: \emph{mock crime}, \emph{best friend}, and \emph{abortion}. In \emph{mock crime}, participants can choose to take or not take the money in the envelope. They were rewarded if they took the money without raising doubts from interviewers. For \emph{best friend} and \emph{abortion}, participants can discuss these topics using true or fake opinions. 
	
	Besides laboratory-controlled scenarios, there are many works focusing on in-the-wild scenarios. For example, \citet{csen2020multimodal} collected videos from public court trials and used trial outcomes to indicate whether the subject was deceptive. \citet{bachenko2008verification} analyzed criminal narratives, interrogations, and legal testimony and provided a method to assess whether a statement is truthful or deceptive. \citet{fornaciari2013automatic} attempted to identify deceptive statements in hearings collected in Italian courts. \citet{perez2015verbal} collected videos from TV shows. The participants were considered to be lying if they gave an opinion about a non-existent movie.
	
	Deception detection mainly uses multimodal clues to identify deceptive behavior. However, such judgment is related to personal habits. Different from deception detection, deception reasoning aims to provide objective evidence for subjective judgment, which has greater value in practical scenarios. For example, during interrogation, these analytical results can provide guidance to the police officer.
	
	\begin{figure*}[t]
		\centering
		\includegraphics[width=\linewidth]{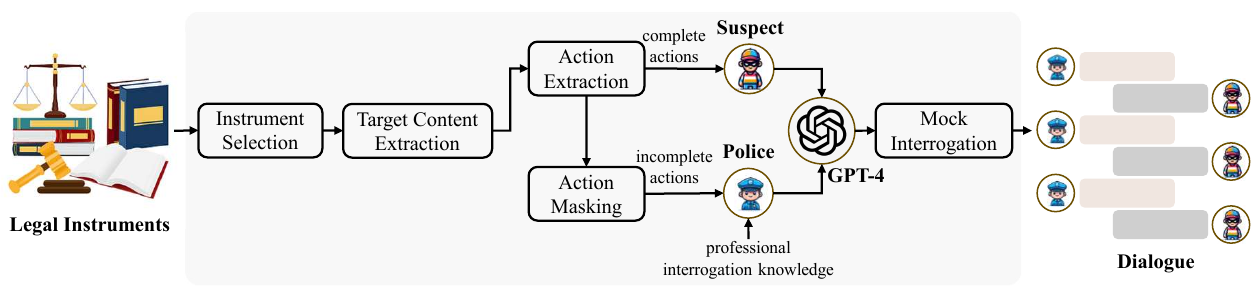}
		\caption{Pipeline of dialogue generation based on legal instruments.}
		\label{Figure1}
	\end{figure*}

	\subsection{Large Language Model}
	Recently, LLMs have shown strong text understanding and generation capabilities, which have been widely used in various tasks and domains. For example, \citet{gan2023large} and \citet{qiu2023smile} explored the promise of LLMs in education and mental health support. \citet{wang2023rolellm} used LLMs to learn character-specific language patterns and behaviors to enhance role-playing realism and interactive experiences. \citet{park2023generative} exploited LLMs to create multiple characters and let them live in a virtual environment. These characters were able to engage in dialogues and spontaneous social activities. Among existing LLMs, GPT-4 shows strong role-playing ability and can generate more human-like behaviors \cite{guo2023evaluating, gui2023challenge}. Therefore, we use GPT-4 to synthesize dialogues for deception reasoning.

	\subsection{Reasoning Performance Evaluation}
	Reasoning is a necessary ability to solve sophisticated problems. For example, mathematical reasoning is the ability to reason about math word problems \cite{mishra2022lila, mishra2022numglue}. Logical reasoning is a cognitive process of applying general rules or principles to reach specific conclusions \citep{flach2013abduction}. In logical reasoning, three elements are usually included: rule, case, and result. These three elements constitute three types of logical reasoning: deductive (\(rule + case \Rightarrow result \)), inductive (\(case + result \Rightarrow rule\)), and abductive (\(result+Rule \Rightarrow case\)). Commonsense reasoning enables computers to understand and apply common knowledge from humans, more effectively simulating human thought processes and decision-making behaviors \citep{storks2019commonsense}.
	
	Existing reasoning datasets mainly use a form of multiple-choice \citep{geva2021did} or open-ended questions \citep{weston2016towards}. For the former, the answer is predefined and the evaluation process is straightforward. For the latter, the model needs to generate the answer, rather than choosing from a given set of options. In our deception reasoning, it is difficult to provide candidate answers and the multiple-choice form may also limit the model's creativity. Therefore, we evaluate this task in the form of open-ended questions.
 
    Previous open-ended questions mainly use the \emph{similarity} between predicted answers and standard answers \cite{yang2018hotpotqa}. Considering the complexity of deception reasoning, this paper proposes a more comprehensive evaluation strategy covering four dimensions: \emph{accuracy}, \emph{completeness}, \emph{logic}, and \emph{depth}. More details can be found in Section \ref{sec:eval}.

	\section{Data Generation}
	\label{sec:data}
	In deception reasoning, we pick a potential lie and analyze why this sentence may be a lie by considering factual inconsistencies and the intent behind it. In this task, we focus on the reasonableness of reasoning rather than the authenticity of deceptive behaviors. Therefore, to reduce the cost of dataset collection, we use GPT-4 to synthesize dialogues containing deceptive behaviors. Specifically, we choose one of the most widely used scenarios in previous works, \emph{mock crime} \cite{derrick2010border, perez2014multimodal}. We ask GPT-4 to simulate the role-playing between a suspect and a police officer. During interrogation, the suspect should deceive the police officer and escape the crime and the police officer should find out the truth.
	
	We first clarify the definition of three important notations: \textbf{\emph{legal instrument}}, \textbf{\emph{target content}}, and \textbf{\emph{action}}. Then, we introduce the data generation process (see Figure \ref{Figure1} for more details), which mainly relies on GPT-3.5 (``gpt-3.5-turbo-0613'') and GPT-4 (``gpt-4-1106-preview'').

	\begin{table*}[t]
		\centering
        \renewcommand\arraystretch{1.2}
		\scriptsize
		\begin{tabular}{p{15.2cm}}
			\hline
			\multicolumn{1}{c}{Legal Instrument} \\
			\hline
			The Tangshan Fengnan District People's Procuratorate accuses: On July 16, 2011, at around 21:00, on the west side of the Pedestrian Street Plaza in Fengnan District, the defendant Zhang, along with Xie Mou (already sentenced), Wang Mou (separate case), and others, demanded the phone number from Feng Mou. After being rejected, they continued to verbally harass. Later, the defendant Zhang and Wang Mou used roller skates, while Xie Mou and others used fists and feet to assault Ma Mou, Tao Mou, Xue Mou, and others who tried to intervene. This resulted in Ma Mou sustaining light injuries, Xue Mou minor injuries, and Tao Mou minor injuries. On the evening of February 11, 2012, at around 19:00, the defendant Zhang, driving a black Santana 3000 sedan (without a license plate), was found at the Lights KTV in Fengnan District, suspected of being involved in the January 31, 2012 case at the Fengnan District Billiard Hall. The incident was immediately reported to the Fengnan District Public Security Bureau, notifying police officer Xue Mou. At the south entrance of Dexin Garden in Fengnan District, when police officer Xue Mou and two colleagues intercepted the defendant Zhang in a car, the defendant Zhang stabbed Xue Mou with a knife and fled, causing minor injuries to Xue Mou. In response to the alleged facts, the public prosecution submitted corresponding evidence. The public prosecution authorities believe that the actions of Defendant Zhang constitute the crimes of xxx and xxx and request sentencing according to the provisions of the Criminal Law of the People's Republic of China xxx and xxx. \\
			\hline
			\multicolumn{1}{c}{Target Content} \\
			\hline
			1. On July 16, 2011, around 21:00, on the west side of the Pedestrian Street Plaza in Fengnan District, the defendant Zhang, along with Xie Mou (already sentenced), Wang Mou (separate case), and others, demanded the phone number from Feng Mou. After being rejected, they continued to verbally harass. Later, the defendant Zhang and Wang Mou used roller skates, while Xie Mou and others used fists and feet to assault Ma Mou, Tao Mou, Xue Mou, and others who tried to intervene. This resulted in Ma Mou sustaining light injuries, Xue Mou minor injuries, and Tao Mou minor injuries. \\
			2. On the evening of February 11, 2012, at around 19:00, the defendant Zhang, driving a black Santana 3000 sedan (without a license plate), was found at the Lights KTV in Fengnan District, suspected of being involved in the January 31, 2012 case at the Fengnan District Billiard Hall. The incident was immediately reported. At the south entrance of Dexin Garden in Fengnan District, the defendant Zhang used a knife to injure Xue Mou and fled, causing minor injuries to Xue Mou. \\
			\hline
			\multicolumn{1}{c}{Complete Actions} \\
			\hline
			1. On July 16, 2011, around 21:00, on the west side of the Pedestrian Street Plaza in Fengnan District, the defendant Zhang, along with Xie Mou and Wang Mou, demanded the phone number from Feng Mou but was refused. \\
			2. On July 16, 2011, the defendant Zhang and Wang Mou used roller skates, while Xie Mou and others used fists and feet to assault Ma Mou, Tao Mou, Xue Mou. This resulted in Ma Mou sustaining light injuries, Xue Mou minor injuries, and Tao Mou minor injuries. \\
			3. On the evening of February 11, 2012, at around 19:00, the defendant Zhang, driving a black Santana 3000 sedan (without a license plate), was found at the Lights KTV in Fengnan District. Someone suspected that he was involved in a previous case and immediately reported it to the Fengnan District Public Security Bureau, notifying police officer Xue Mou. \\
			4. On February 11, 2012, at the south entrance of Dexin Garden in Fengnan District, the defendant Zhang used a knife to injure Xue Mou and fled. This attack caused minor injuries to Xue Mou. \\
			\hline
			\multicolumn{1}{c}{Incomplete Actions} \\
			\hline
			1. At an unknown time, on the west side of the Pedestrian Street Plaza in Fengnan District, the defendant Zhang, along with Xie and Wang, demanded Feng's phone number, but was refused. \\
			
			2. On July 16, 2011, the defendant Zhang and Wang, using unknown tools, along with Xie and others using fists and feet, assaulted Ma, Tao, Xue. This assault resulted in Ma suffering minor injuries, Xue having minor injuries, and Tao having minor injuries. \\
			
			3. On February 11, 2012, around 7:00 PM, the defendant Zhang drove a black Santana 3000 sedan (without a license plate), and at an unknown location, was found by someone who immediately reported it to Fengnan District Public Security Bureau police officer Xue, suspecting involvement in a previous case. \\
			
			4. On February 11, 2012, at the south entrance of Dexin Garden in Fengnan District, the defendant Zhang used unknown tools to injure Xue and then fled. This attack caused Xue to suffer minor injuries. \\
			
			\hline
		\end{tabular}
		\caption{Examples of the legal instrument, target content, and action.}
		\label{Table1}
	\end{table*}

	\subsection{Notation Definition}
	In this paper, we ask GPT-4 to conduct mock interrogation around the crime facts between a suspect and a police officer. To obtain crime facts, we turn our attention to \textbf{\emph{legal instruments}}, which include but are not limited to, details of the prosecution's charges, descriptions of the defendant's criminal behavior, arrests, the evidence presented, explicit charges, and stages of the judicial process.
	
	To mimic real interrogation, the suspect should know the complete crime facts while the police officer should miss some details. However, \emph{legal instruments} contain contents that can reduce uncertainty during interrogation, such as explicit charges and convictions. Hence, in \emph{legal instruments}, we only select the \textbf{\emph{target content}}, which denotes a series of behaviors involving multiple people, places, and times. The \emph{target content} contains multiple \textbf{\emph{actions}}, where an \emph{action} refers to a continuous and specific behavior performed by subjects within a period of time. Table \ref{Table1} provides examples of the \emph{legal instrument}, \emph{target content}, and \emph{action}.
	
	\subsection{Legal Instrument Selection}
	\label{sec:3-2}
	CAIL2018 \citep{xiao2018cail2018} encompasses 2.68 million criminal law documents, spanning 202 types of charges and 183 legal provisions. In this dataset, legal instruments are written by legal experts, with rigorous wording and standardized forms. These high-quality legal instruments bring great convenience to our work.
	
\begin{figure}[t]
		\centering
		\includegraphics[width=0.9\linewidth]{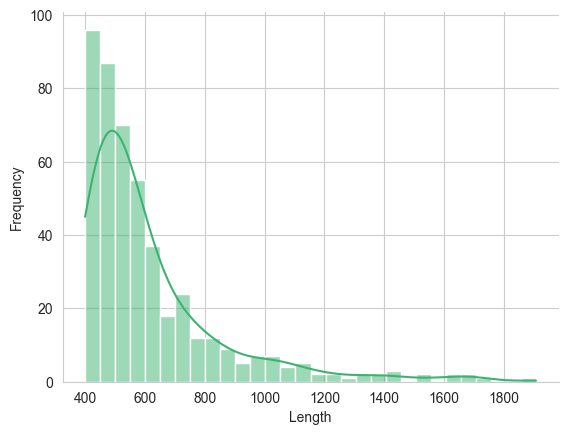}
		\caption{Distribution of lengths after selection (the length refers to the number of Chinese characters).}
		\label{Figure2}
	\end{figure}

	Proper legal instruments are important for dialogue generation. On the one hand, short legal instruments contain insufficient content, leading to unclear descriptions of details and generating low-quality dialogues. On the other hand, long legal instruments may contain complex crime facts, increasing the difficulty of dialogue generation. Therefore, we select legal instruments with a length ranging from 400 to 2,000. The length distribution after selection is shown in Figure \ref{Figure2}, where the length refers to the number of Chinese characters.

	\subsection{Target Content and Action Extraction}
	\label{sec:target_extraction}
	In this section, we aim to extract the \emph{target content} from \emph{legal instruments} and further disassemble it into multiple \emph{actions}. Specifically, we rely on GPT-4 and adopt a two-stage strategy to achieve this goal. In the first stage, we extract the \emph{target content} from \emph{legal instruments}; in the second stage, we disassemble it into multiple \emph{actions}. To achieve better performance, each stage uses one-shot and chain-of-through prompts \cite{wei2022chain}. 
 
    In this paper, we also analyze the performance of the one-stage strategy, i.e., merging \emph{target content} and \emph{action} extraction into one stage. Experimental results demonstrate that the two-stage strategy is more effective than the one-stage strategy. Meanwhile, GPT-4 performs better than GPT-3.5. More details can be found in Section \ref{sec:onestage}.

	\subsection{Incomplete Action Generation}
	\label{sec:mask}
	During the interrogation, the police officer may not have complete crime facts and try to find missing parts from the suspect. To mimic this process, we generate incomplete actions for the police officer. 
	
	An action mainly involves the following seven items: (1) \emph{agent} is the entity that performs the action; (2) \emph{patient} is the entity affected by the action; (3) \emph{instrument} is the object used to perform the action; (4) \emph{goal} is the direction or destination of the action; (5) \emph{source} is the place where the action originates; (6) \emph{time} is the time when the action occurs; (7) \emph{location} is the place where the action occurs.

	\begin{figure}[t]
		\centering
		\includegraphics[width=0.9\linewidth]{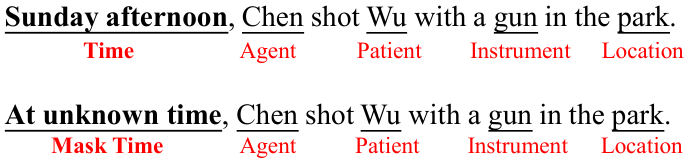}
		\caption{Example of time masking process.}
		\label{Figure3}
	\end{figure}

	\begin{figure*}[t]
		\centering
		\includegraphics[width=\linewidth]{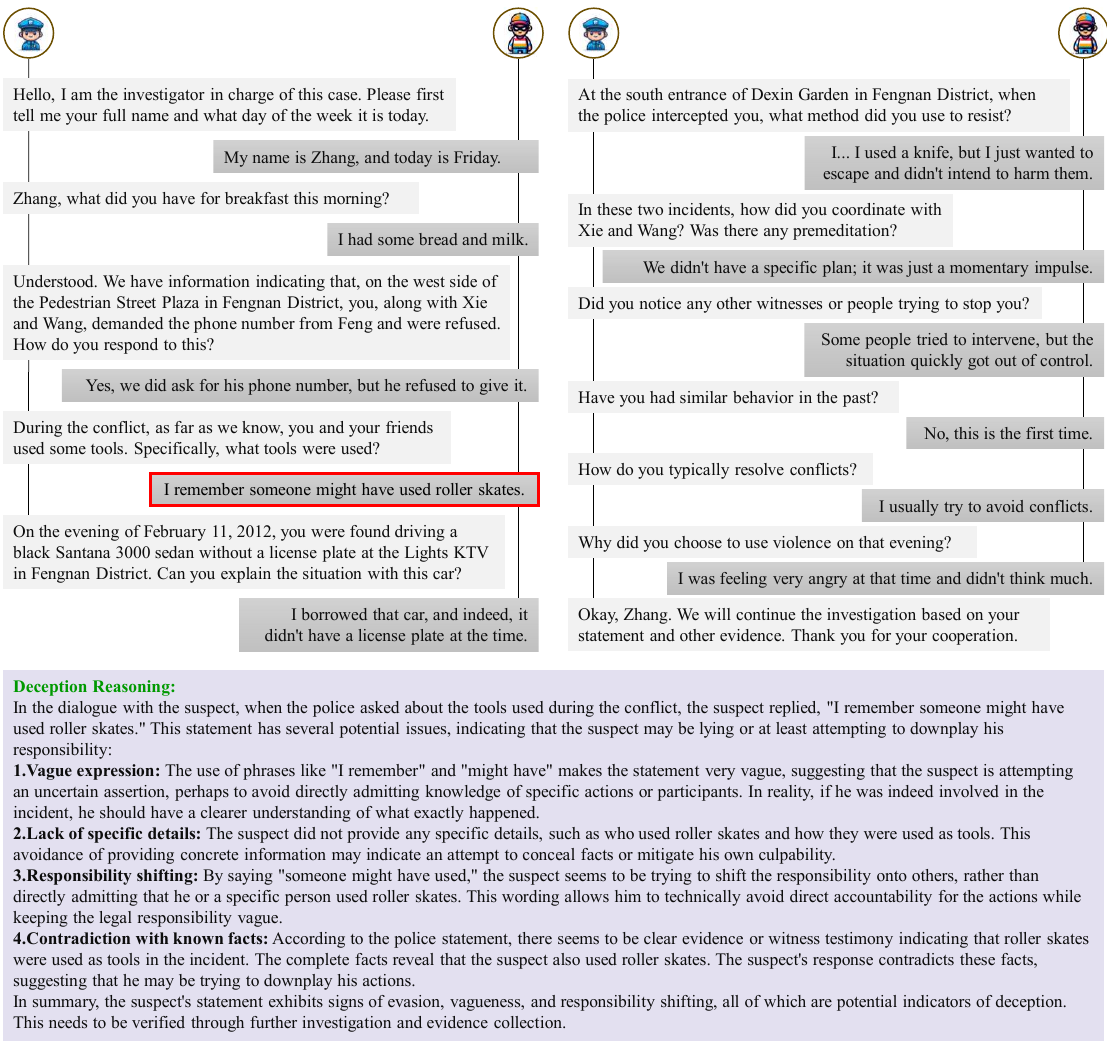}
		\caption{Generated dialogue, potential lie (in the red box), and reasoning results using examples in Table \ref{Table1}.}
		\label{Figure4}
	\end{figure*}
	
	To generate incomplete actions, we randomly mask an item in the action. Specifically, we replace the \emph{agent} and \emph{patient} with unknown people, the \emph{instrument} with unknown tool, the \emph{location} with unknown place, and the specific \emph{time} with unknown time. We provide an example in Figure \ref{Figure3}. Table \ref{Table1} shows the generated incomplete actions. This masking process is also realized by GPT-4.

	\begin{table*}[t]
		\centering
		\small
		\begin{tabular}{c|c}
			\hline
			{Metric} & Value \\
			\hline
			\# of dialogues & 191 \\
			max/min/avg \# of turns per dialogue & 54/23/34.93 \\
			max/min/avg \# of words per utterance & 180/2/19.3 \\
			max/min/avg \# of words per police's utterance & 180/7/24.23 \\
			max/min/avg \# of words per suspect's utterance & 99/7/20.77  \\
			max/min/avg police word count divided by suspect word count per turn &  9.0/0.17/1.27\\
			\hline
		\end{tabular}
		\caption{Statistics of our generated deception dataset.}
		\label{Table2}
	\end{table*}

	\begin{figure*}[t]
		\begin{center}
			\subfloat[target content length]{
				\label{Figure5-1}
				\centering
				\includegraphics[width=0.32\linewidth]{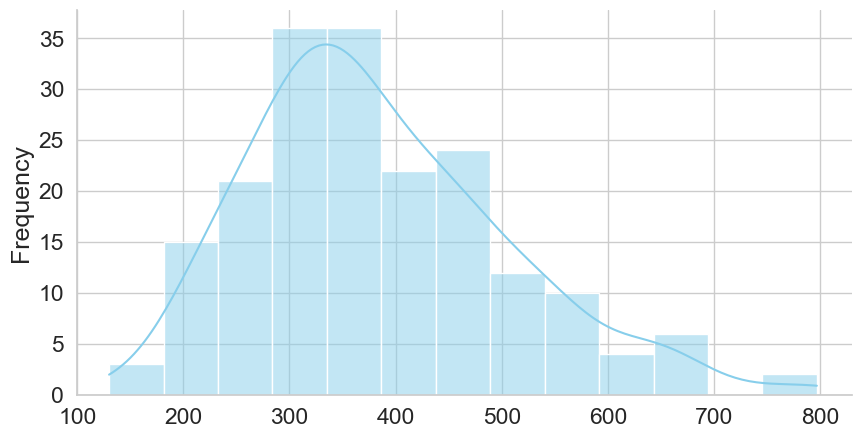}}
			\subfloat[number of actions]{
				\label{Figure5-2}
				\centering
				\includegraphics[width=0.32\linewidth]{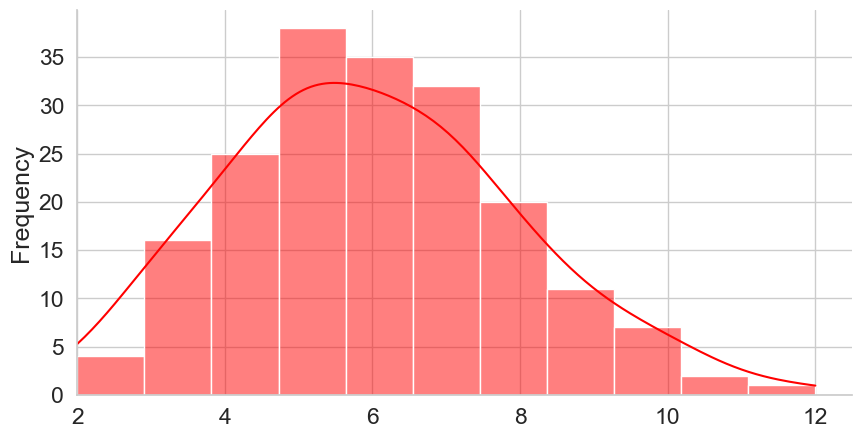}}
			\subfloat[dialogue turns]{
				\label{Figure5-3}
				\centering
				\includegraphics[width=0.32\linewidth]{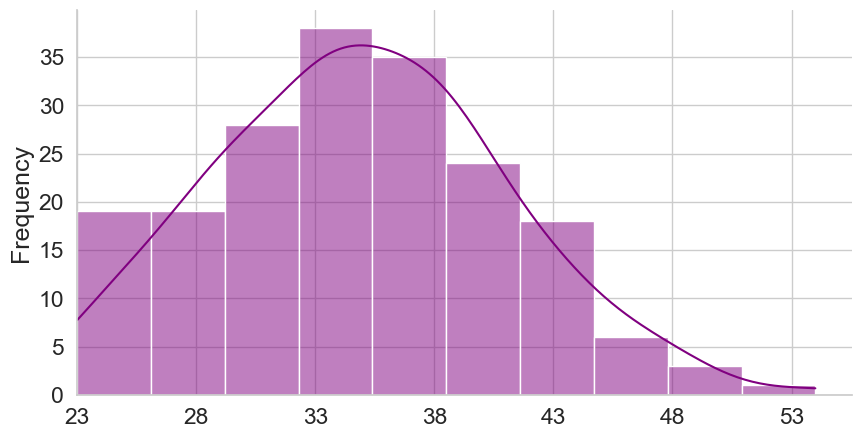}}
		\end{center}
		\caption{Distribution of target content length, number of actions, and dialogue turns.}
		\label{Figure5}
	\end{figure*}

	\subsection{Mock Interrogation}
	\label{sec:3-5}
	We simulate the interrogation process between the suspect and the police officer. To enhance authenticity, complete and incomplete actions serve as the information held by the suspect and the police officer, respectively. To enhance the professionalism of the police officer, we further provide him with additional interrogation techniques. Figure \ref{Figure4} provides the generated dialogue for examples in Table \ref{Table1}. Specifically, we require the police officer to ask some typical questions \citep{leo1994police}:

	\begin{table*}[t]
		\centering
		\renewcommand\tabcolsep{8.8pt}
		\renewcommand\arraystretch{0.9}
		\small
		\begin{tabular}{c|ccccc|ccccc}
			\hline
			\multirow{2}{*}{Model} & \multicolumn{5}{c|}{Automatic Evaluation Results} & \multicolumn{5}{c}{Manual Evaluation Results} \\
			& Acc. & Com. &Log. & Dep. & Sum & Acc. & Com. &Log. & Dep. & Sum\\
			\hline
			ChatGLM2-6B			&4.00 &3.56 &4.33 &3.44	&15.33 &5.12 &5.00 &4.90 &4.70 &19.72 \\
			WizardLM-13B	 	&5.20 &4.87 &6.00 &4.38	&20.45 &5.61 &5.30 &5.41 &5.14 &21.46 \\
			Baichuan2-13B	 	&5.24 &5.00 &6.25 &4.62	&21.11 &4.56 &4.36 &4.49 &4.16 &17.57 \\
			ERINE3.5		 	&5.40 &5.00 &6.10 &5.10	&21.60 &5.71 &5.81 &5.71 &5.13 &22.36 \\
			Qwen-14B		 	&6.00 &5.60 &6.70 &5.20	&23.50 &5.91 &5.80 &5.42 &5.40 &22.53 \\
			Claude3-Haiku	 	&6.33 &5.89 &6.89 &5.33	&24.44 &6.36 &6.11 &5.94 &5.70 &24.11 \\
			GPT-3.5			 	&6.00 &5.87 &6.87 &5.75	&24.49 &6.80 &6.58 &6.53 &6.18 &26.09 \\
			ERINE4.0		 	&6.60 &6.30 &7.30 &5.80	&26.00 &6.95 &6.78 &6.99 &6.81 &27.53 \\
			GLM-4-9B		 	&6.67 &6.44 &7.33 &6.33	&26.77 &7.56 &7.54 &7.59 &7.55 &30.24 \\
			Gemini-1.5-Pro	 	&6.11 &6.89 &7.67 &6.56	&27.23 &7.56 &7.51 &7.37 &7.23 &29.67 \\
			Qwen2-7B		 	&6.56 &6.72 &7.72 &6.39	&27.39 &7.41 &7.49 &7.48 &7.41 &29.79 \\
			\rowcolor{lightgray}
			PCC scores          &0.81 &0.87 &0.80 &0.89 &0.86 &0.81 &0.87 &0.80 &0.89 &0.86 \\   
			\hline       
		\end{tabular}
		\caption{Main results of different LLMs on four evaluation metrics. We show both automatic and manual evaluation results, and the last row reports the PCC scores between them.}
		\label{Table3}
	\end{table*}

	\begin{itemize}
		\item Control questions: These questions are used to establish a baseline response from the interrogatee. Generally, the interrogatee is honest with these questions. For example, what is your name? What day of the week is it today? Answers to these questions should be truthful so that they can be compared with answers to subsequent questions.
		
		\item Relevant questions: They are related to the core of the crime and are often questions to get to the truth. For example, were you involved in an event at a certain time and place? How did you do this? The answers to these questions are the focus of the interrogation.
		
		\item Comparison questions: These questions are similar to control questions, but they are usually designed to be more challenging to show a distinct physical or psychological response. These questions should be answered in the affirmative. For example, have you ever done anything dishonest? Do you lie often?
		
		\item Neutral questions: These questions are often used to relieve tension or provide an opportunity for the interrogatee to relax. They are not related to the subject of the interrogation. For example, what did you have for breakfast this morning? What are your hobbies?
		
		\item Randomness and variability: Interrogators usually randomize the order of questions to avoid forming a fixed pattern, thereby reducing the chances that the interrogatee will be able to prepare for or adapt to a particular type of questioning, but neutral and control questions often come first in interrogation.
		
	\end{itemize}
	
	In this section, we propose two strategies for dialogue generation: (1) we use two GPT-4s playing two roles; (2) we use one GPT-4 to directly generate a multi-round dialogue between two roles. For the first strategy, the output gradually spirals out of control as the dialogue progresses, resulting in a significant drop in quality at the end of the dialogue. Therefore, we turn our attention to the second strategy. We find this strategy can maintain the logic and coherence of the dialogue.

	\subsection{Complete Pipeline and Data Statistics}
	In this section, we summarize the data construction pipeline. Specifically, we first remove legal documents with inappropriate length (see Section \ref{sec:3-2}). Then, we randomly sample legal documents and synthesize dialogues (see Section \ref{sec:target_extraction}$\sim$Section \ref{sec:3-5}). After that, we manually select a potential lie that is more representative of humans and analyze why this sentence may be a lie by considering factual inconsistencies and the intent behind it. Figure \ref{Figure4} gives an example to illustrate this process. To ensure the quality of data, we further perform post-filtering to remove some dialogues that contain unnatural parts or without potential lies. Finally, we generate 191 dialogues.
	
	Dataset statistics are summarized in Table \ref{Table2}. We observe that the average number of turns per dialogue is 34.93, which is sufficient for a short interrogation. In Figure \ref{Figure5}, we also provide the distribution of target content length, number of actions, and dialogue turns. Meanwhile, we analyze the cost of data collection. On average, we spend less than \$2 per dialogue. Compared with existing datasets, subject recruitment and data annotation often require a lot of money, and the cost varies from country to country. But in our country, it costs more than \$2 per dialogue. Therefore, this paper provides a cheaper way to collect data.
	
	It should be noted that deception reasoning has some similarities with misinformation detection. However, there are also certain differences between them. As shown in Figure \ref{Figure4}, misinformation detection is one aspect of deception reasoning. Besides this, deception reasoning also focuses on other aspects, such as ambiguous expressions.
	
	\section{Deception Reasoning Evaluation}
	\label{sec:eval}
	In this section, we first define evaluation metrics and evaluators. Then, we assess different LLMs and report evaluation results. After that, we prove the naturalness of synthetic dialogues. Finally, we conduct an ablation study and reveal the rationality of our target content and action extraction strategy. This section mainly uses GPT-3.5 (``gpt-3.5-turbo-0613'') and GPT-4 (``gpt-4-1106-preview'').

	\subsection{Evaluation Metrics}
    In deception reasoning, we need to figure out why a sentence may be a lie by considering factual inconsistencies and the intent behind it. To provide a more comprehensive evaluation, we propose four metrics for deception reasoning, whose core definitions are provided below:
	
	\begin{itemize}
		\item Accuracy: It is used to check whether the reasoning is consistent with the basic facts. If the reasoning is based on the facts, the model should receive a high score in this dimension.
		
		\item Completeness: It is used to evaluate whether the model takes into account all details. A good model should be comprehensive and not miss any key information.
		
		\item Logic: It is used to evaluate whether the reasoning is logically coherent and well organized. The model is required to have common sense and world knowledge, with deductive, inductive, abductive, and other reasoning abilities. If the reasoning is logically confused or contradictory, the model should receive a low score in this dimension.
		
		\item Depth: It is used to evaluate whether a model provides an in-depth analysis or only scratches the surface. This metric is different from completeness. Some reasoning merely restates facts and gives a conclusion, which can be complete but not deep. High-quality reasoning should be able to dig deeper into the reasons and motivations behind it.
		
	\end{itemize}

	During the evaluation, we use more detailed definitions for each metric, as well as the meaning of each score for each metric in our prompts. More details can be found in Tables \ref{Table6}$\sim$\ref{Table9}.

	\subsection{Evaluator}
	We conduct both automatic and manual evaluations. Considering that researchers \cite{zheng2023judging, lian2023explainable} have proven the consistency between GPT-4 and human assessments, we directly use GPT-4 for automatic evaluator. Meanwhile, considering that using one LLM to evaluate other LLMs may lead to bias issues and the	possibility of overfitting, we further hire 8 annotators and perform manual evaluation. Each annotator is paid \$14 per hour, which is relatively high in our country.

	\subsection{Main Results}
	This section evaluates the deception reasoning performance of different LLMs. Besides mainstream LLMs such as WizardLM-13B \cite{xu2023wizardlm}, we also select LLMs that perform well in Chinese. During inference, we input basic facts, synthetic dialogue, and potential lies, and ask LLMs to analyze why this sentence might be a lie. In both automatic and manual evaluation, we use the prompts in Tables \ref{Table6}$\sim$\ref{Table9} and experimental results are shown in Table \ref{Table3}. We observe that existing LLMs can deal with deception reasoning to some extent. Meanwhile, we can see the progress of Chinese LLMs in reasoning ability. For example, Qwen2 is better than Qwen and ERINE4.0 is better than ERINE3.5. Furthermore, Table \ref{Table3} shows the PCC scores between automatic and manual evaluation results. We observe that manual evaluation results have relatively high similarities with automatic evaluation results, proving the reliability of our automatic evaluation strategy.

	Table \ref{Table4} shows the inference cost per sample for each LLM. For closed-source models provided by OpenAI, Google, etc., we calculate the inference cost based on the number of tokens and the price per token. For open-source models such as GLM-4-9B and Qwen2-7B, we calculate the inference cost based on the model inference time and the daily price of the machine usage. Specifically, we use Azure Standard\_NC12s\_v3 (equipped with 2 V100 GPUs) based on the pay-as-you-go pricing in December 2023. Although these costs are not accurate due to price changes, they provide a rough estimate of the inference cost. We find that for open-source LLMs, large models are often expensive due to their long inference time. For close-source LLMs, Gemini-1.5-Pro is cheaper than GPT-3.5.

	\begin{table}[t]
		\centering
		\small
            \renewcommand\arraystretch{0.9}
		\begin{tabular}{c|c}
			\hline
			{Model} & Cost ($\times 10^{-3} \$ $) \\
			\hline
			ChatGLM2-6B		 &1.3	\\
			WizardLM-13B	 &3.6	\\
			Baichuan2-13B	 &2.1   \\
			ERINE3.5		 &0.1	\\
			Qwen-14B		 &2.2	\\
			Claude3-Haiku	 &0.9	\\
			GPT-3.5			 &4.2	\\
			ERINE4.0		 &3.6	\\
			GLM-4-9B		 &2.8	\\
			Gemini-1.5-Pro	 &0.7	\\
			Qwen2-7B		 &1.8	\\
			\hline
		\end{tabular}
		\caption{Inference cost per sample for different LLMs.}
		\label{Table4}
	\end{table}

	\subsection{Dialogue Naturalness Evaluation}
	To test the naturalness of our synthetic dialogue, we use the prompt in Table \ref{Table10} and conduct both automatic and manual evaluations. 
	
	In the automatic evaluation process, considering that we use GPT-4 to generate dialogues, we choose another Claude3-Haiku for evaluation. Specifically, we randomly select 10 real dialogues from a dialogue dataset IEMOCAP \cite{busso2008iemocap} and 10 synthetic dialogues from our dataset. The average score of real dialogue can reach 4.00 and the average score of synthetic dialogue can reach 3.88.
	
	In the manual evaluation process, we hire eight annotators and ask them to score the naturalness. We observe that the average score of synthetic dialogue can reach 3.70, close to the automatic evaluation results. All these results reflect the naturalness of our synthetic dialogues and the reliability of our automatic evaluation strategy.

	\subsection{Ablation Study}
	\label{sec:onestage}
	This paper uses a two-stage strategy and GPT-4 for target content and action extraction (see Section \ref{sec:target_extraction}). In this section, we compare the performance between one-stage and two-stage strategies, as well as GPT-3.5 and GPT-4. During evaluation, we hire one annotator and randomly select 100 samples. For target content extraction, we define a metric called \emph{target accuracy}. If the system extracts non-target content from legal instruments, it will have a low score in this metric. For action extraction, we define a metric called \emph{action complexity}. This metric is related to inappropriate action decomposition. Take the complete actions in Table \ref{Table1} as	an example. These actions are well-decomposed. But if we merge two actions into one action, this decomposition process is inappropriate, leading to an increase in \emph{action complexity}. Therefore, a good model should have high \emph{target accuracy} and low \emph{action complexity}. Experimental results of different strategies are shown in Table \ref{Table5}.
	
	From this table, we observe that the two-stage strategy achieves better performance than the one-stage strategy. The reason lies in that if we merge target content and action extraction into one stage, it increases the task difficulty, making it more likely that the output does not meet the requirements.
	
	Meanwhile, GPT-4 can achieve better performance than GPT-3.5. Target content and action extraction require the model to understand not only the literal meaning of the text but also its structure and semantic content. Since GPT-4 can achieve better performance than GPT-3.5 in text understanding, it can also achieve better performance in target content and action extraction.

	\begin{table}[t]
		\centering
		\small
		\begin{tabular}{l|cc}
			\hline
			Strategy & {Target $(\uparrow)$} & {Action $(\downarrow)$}\\
			\hline
			one-stage + GPT-3.5 &47 &36\\
			two-stage + GPT-3.5 &83 &9 \\
			\hline
			one-stage + GPT-4   &69 &2 \\
			two-stage + GPT-4   &98  &0 \\
			\hline
		\end{tabular}
		\caption{Performance comparison of different strategies for target content and action extraction.}
		\label{Table5}
	\end{table}

	\section{Conclusions}
	\label{sec:conclusion}
	This paper extends deception detection to deception reasoning, further providing objective evidence to support subjective judgment. To facilitate subsequent research, we build a dataset, define evaluation metrics, and open-source data and code. Then, we reveal the performance of mainstream LLMs and show the progress of Chinese LLMs in reasoning ability. Meanwhile, we prove the rationality of our dataset construction strategy and the naturalness of our synthetic dialogues. This task can also serve as a reasoning benchmark for current LLMs.

	\section*{Limitations}
	\label{sec:limitations}
	There are several limitations that can be addressed in future research. First, our deception dataset relies on GPT-4, which requires API call costs. Therefore, we only select a part of legal instruments from CAIL2018 instead of using the entire dataset. Future research will consider using all legal instruments for dialogue generation. Secondly, this paper evaluates the performance of mainstream LLMs but does not cover all LLMs. In the future, we will expand the evaluation scope. Thirdly, we focus on the reasonableness of reasoning rather than the authenticity of deceptive behaviors. Therefore, to reduce the cost of data collection, this paper mainly uses synthetic dialogues. In the future, we will also do some experiments on real interrogation dialogues. Fourthly, video generation has become increasingly popular. In the future, we will synthesize multimodal data and expand text-based deception reasoning to multimodal deception reasoning.

	\section*{Societal Impacts}
	\label{sec:societal_impacts}
	This paper uses legal instruments for dataset construction. On the one hand, legal instruments may provide guidance to criminals. But on the other hand, legal instruments can also remind people not to commit crimes. This paper has similar potential societal impacts as legal instruments. Although our research revolves around deception, our main goal is to detect deception and provide evidence to support the judgment. This tool is of great significance for the police to improve integration efficiency and strengthen social security.

	\bibliography{mybib}
	
	\appendix
	
	\section{Metric Calculation}
	In deception reasoning, we define four evaluation metrics: \emph{accuracy}, \emph{completeness}, \emph{logic}, and \emph{depth}. In Tables \ref{Table6}$\sim$\ref{Table9}, we provide detailed definitions for each metric, as well as the meaning of each score.

	\section{Dialogue Naturalness Evaluation}
	We rank the dialogue naturalness using five scores. In Table \ref{Table10}, we provide the meaning of each score.

	\begin{table*}[t]
		\centering
		\small
		\begin{tabular}{p{15cm}}
			\hline
			
			We provide facts, a dialogue, and a potential lie. Meanwhile, we provide a model's analysis results of whether this sentence might be a lie. Please evaluate whether the model's inference aligns with known facts. If an inference is closer to the real situation or facts, it should receive a higher score in this dimension. \\
			
			\textbf{0-2 points (Very low accuracy):} Most of the arguments do not align with known facts, with only a small part possibly slightly related. Displays a severe misunderstanding of the facts or selective ignorance.   \\
			
			\textbf{3-4 points (Low accuracy):} Some of the arguments align with the facts, but most of the content is still inaccurate or misleading. There is an attempt to use correct facts, but they are handled improperly or misunderstood. \\
			
			\textbf{5-6 points (Moderate accuracy):} There is some degree of consistency between the arguments and the facts, but there are still noticeable inaccuracies. Displays a basic understanding of the facts, but lacks thorough or detailed consideration. \\
			
			\textbf{7-8 points (High accuracy):} Most of the arguments align with the facts, with only a few details or aspects showing deviations. Shows a good understanding of the facts and correct application, but there is still room for improvement.  \\
			
			\textbf{9-10 points (Very accurate):} All or almost all of the arguments strictly align with known facts. Accurately and fully understands and applies the facts, with no obvious errors or omissions. \\
			
			\hline
		\end{tabular}
		\caption{Prompt for evaluating the accuracy.}
		\label{Table6}
	\end{table*}

	\begin{table*}[t]
		\centering
		\small
		\begin{tabular}{p{15cm}}
			\hline
			
			We provide facts, a dialogue, and a potential lie. Meanwhile, we provide a model's analysis results of whether this sentence might be a lie. Please evaluate whether the model has considered all relevant information and details. A good inference should be comprehensive, without omitting any key points. I will provide a "possible answer" which can be considered a reliable answer scoring 8 or above, as a reference for completeness scoring. \\
			
			\textbf{0-2 points (Very low completeness):} The inference is extremely one-sided or missing significant content, with almost no resemblance to the "possible answer." Ignores key aspects of the problem, possibly only scratching the surface. Lacks basic understanding of the problem, with results far from the "possible answer." \\
			
			\textbf{3-5 points (Moderate completeness):} The inference includes some key aspects of the problem but still falls far short of the "possible answer."	There is an attempt to address the problem comprehensively, but some important aspects or details are overlooked. The inference has some resemblance to the "possible answer," but there are still clear omissions or misunderstandings. \\
			
			\textbf{6-7 points (High completeness):} The inference is fairly comprehensive, covering most key aspects and closely aligning with the "possible answer." Able to understand and respond well to the core requirements of the problem, though there may still be some omissions in details. The inference has a high degree of similarity to the "possible answer," but there is still room for improvement. \\
			
			\textbf{8 points (Very high completeness):} The inference is very comprehensive, covering all key aspects of the problem and highly consistent with the "possible answer." Demonstrates a deep understanding of the problem, responding accurately and thoroughly to all aspects. The result shows depth and detail, with almost no omissions or misunderstandings. \\
			
			\textbf{9-10 points (Exceeds completeness):}	Not only highly consistent with the "possible answer," but also offers innovation or further depth. Provides additional insights. \\
			\hline
		\end{tabular}
		\caption{Prompt for evaluating the completeness.}
		\label{Table7}
	\end{table*}

	\begin{table*}[!t]
		\centering
		\small
		\begin{tabular}{p{15cm}}
			\hline
			
			We provide facts, a dialogue, and a potential lie. Meanwhile, we provide a model's analysis results of whether this sentence might be a lie. Please evaluate whether the inference is logically coherent and well-structured. If the model provides an inference that is disorganized or self-contradictory, it should score lower in this dimension. \\
			
			\textbf{0-2 points (Low level):} The inference has almost no logical coherence, possibly entirely based on guesswork or conjecture. There is no clear connection between evidence and conclusions, or no evidence is used at all. The reasoning process is chaotic and lacks organization, possibly deviating completely from the core of the problem. \\
			
			\textbf{3-5 points (Moderate level):} The inference has some logical coherence, but there may be noticeable logical gaps or errors. Some relevant evidence is used, but the application of evidence is either inappropriate or insufficient. The reasoning process, while somewhat structured, may lack in-depth analysis in key areas. \\
			
			\textbf{6-8 points (Good level):} The inference has strong logical coherence, with few logical gaps. Evidence is used appropriately and sufficiently to support the conclusion. The reasoning process is clear and well-organized, with in-depth analysis of key parts of the problem. \\
			
			\textbf{9-10 points (Excellent level):} The inference is highly logical, with almost no logical gaps. Evidence is used extremely appropriately and sufficiently, strongly supporting the conclusion. The reasoning process is very clear and well-organized, with in-depth exploration of various aspects of the problem, possibly providing new insights or solutions. \\
			\hline
		\end{tabular}
		\caption{Prompt for evaluating the logic.}
		\label{Table8}
	\end{table*}

	\begin{table*}[!t]
		\centering
		\small
		\begin{tabular}{p{15cm}}
			\hline
			We provide facts, a dialogue, and a potential lie. Meanwhile, we provide a model's analysis results of whether this sentence might be a lie. Please evaluate whether there has been deep thought and analysis or if it remains superficial. A high-quality inference should be able to deeply explore underlying reasons and motivations. \\
			
			\textbf{0-2 points (Superficial thinking):} The inference is superficial, staying only at the surface level. Lacks exploration of underlying reasons and motivations, ignoring the complexity and deeper factors of the problem. The conclusion may be overly simple and direct, not showing multi-angle or in-depth consideration of the problem. \\
			
			\textbf{3-5 points (Basic depth):} The inference shows some depth, but still not comprehensive or deep enough. While there is an attempt to explore underlying reasons and motivations, the analysis still appears shallow or incomplete. The reasoning process has some logical coherence, but lacks depth and complexity. \\
			
			\textbf{6-8 points (Good depth):} The inference shows good depth, exploring various aspects of the problem fairly comprehensively. There is in-depth analysis of underlying reasons and motivations, providing deep insights. Although the depth is high, there may still be room for further exploration in some areas. \\
			
			\textbf{9-10 points (Extremely deep):} The inference is extremely deep and comprehensive, uncovering the core and deeper factors of the problem. The analysis of underlying reasons and motivations is profound, offering unique and deep insights. The reasoning process displays a high level of logical coherence and depth, exploring the problem from multiple angles, showing high-quality thinking. \\
			
			\hline
		\end{tabular}
		\caption{Prompt for evaluating the depth.}
		\label{Table9}
	\end{table*}
	
	\begin{table*}[t]
		\centering
		\small
		\begin{tabular}{p{15cm}}
			\hline
			Now you need to rate a conversation. Please ignore its format and focus on the content. The more the conversation resembles a real dialogue, the higher the score. The maximum score is five points. The rating criteria are as follows: \\
			
			\textbf{1 point:} very unnatural. The conversation appears very stiff and unnatural, possibly containing numerous grammar errors, incoherent sentences, or content that is completely unrelated to the context. This type of conversation is difficult to understand and gives off a mechanical or robotic feel, lacking the natural fluency of human communication. \\
			
			\textbf{2 points:} somewhat unnatural. Although the conversation conveys basic information, it still seems somewhat unnatural. There may be some linguistic or logical inconsistencies that make the conversation lack the smoothness of natural communication. The conversation may occasionally contain content that is unrelated to the context, requiring further improvement to enhance its naturalness. \\
			
			\textbf{3 points:} moderately natural. The conversation is somewhat fluent but still has some issues. There may be some lack of coherence in some places, or occasional unnatural expressions. The conversation can generally stay on topic but still has room for improvement to better simulate natural language communication. \\
			
			\textbf{4 points:} fairly natural. The conversation is generally fluent and can convey meaning and emotions well. Although there may be occasional minor unnatural aspects, overall, it closely resembles real human dialogue. The conversation is coherent, able to closely follow the topic, and demonstrates good adaptability and understanding. \\
			
			\textbf{5 points:} very natural. The conversation is extremely fluent and natural as if it were a real interaction with a person. There are no language or logical inconsistencies throughout the conversation, maintaining consistency and relevance to the topic. The expression is precise, and adaptable, closely simulating human communication habits and emotional responses, giving a very authentic and comfortable feeling. \\
			\hline
		\end{tabular}
		\caption{Prompt for evaluating the dialogue naturalness.}
		\label{Table10}
	\end{table*}

\end{document}